\documentclass[lettersize,journal]{IEEEtran}

\usepackage{amsmath,amsfonts}
\usepackage{algorithm}
\usepackage{algpseudocode}
\usepackage{subfigure}
\makeatletter
\renewcommand{\@thesubfigure}{\hskip\subfiglabelskip}
\makeatother
\usepackage{array}
\usepackage[caption=false,font=normalsize,labelfont=sf,textfont=sf]{subfig}
\usepackage{textcomp}
\usepackage{stfloats}
\usepackage{multirow}
\usepackage{enumitem}
\setdescription{labelsep=\textwidth}
\usepackage{url}
\usepackage{verbatim}
\usepackage{graphicx}
\usepackage{cite}
\usepackage{amsmath,amssymb}

\title{Functional Optimization Reinforcement Learning for Real-Time Bidding}
\begin{document}
\author{Yining Lu\textsuperscript{*}, Changjie Lu\textsuperscript{*},  Naina Bandyopadhyay, Manoj Kumar, Gaurav Gupta
\thanks{* indicates equal contribution}
\thanks{Yining Lu, Changjie Lu, and Gaurav Gupta are with the School of Mathematics, College of Science and Technology, Wenzhou-Kean University, Wenzhou 325060, China(e-mail: yiningl@kean.edu, lucha@kean.edu, ggupta@kean.edu)}
\thanks{Naina Bandyopadhyay is with the Findability Sciences Private Limited, Mumbai 400057, India(e-mail: bnaina@findabilitysciences.com )}
\thanks{Manoj Kumar is with the Department of Computer Science, Babasaheb Bhimrao Ambedkar University, Lucknow 226025, India}
}



\maketitle



\begin{abstract}
Real-time bidding is the new paradigm of programmatic advertising. An advertiser wants to make the intelligent choice of utilizing a \textbf{Demand-Side Platform} to improve the performance of their ad campaigns. Existing approaches are struggling to provide a satisfactory solution for bidding optimization due to stochastic bidding behavior. In this paper, we proposed a multi-agent reinforcement learning architecture for RTB with functional optimization. We designed four agents bidding environment: three Lagrange-multiplier based
functional optimization agents and one baseline agent (without any attribute of functional optimization)
First, numerous attributes have been assigned to each agent, including biased or unbiased win probability, Lagrange multiplier, and click-through rate. In order to evaluate the proposed RTB strategy's performance, we demonstrate the results on ten sequential simulated auction campaigns. The results show  that agents with functional actions and rewards had the most significant average winning rate and winning surplus, given biased and unbiased winning information respectively. The experimental evaluations show that our approach  significantly improve the campaign's efficacy and profitability.
\end{abstract}


\section{Introduction}
\begin{figure}[t]
    \centering
    \includegraphics[width = 3.5in]{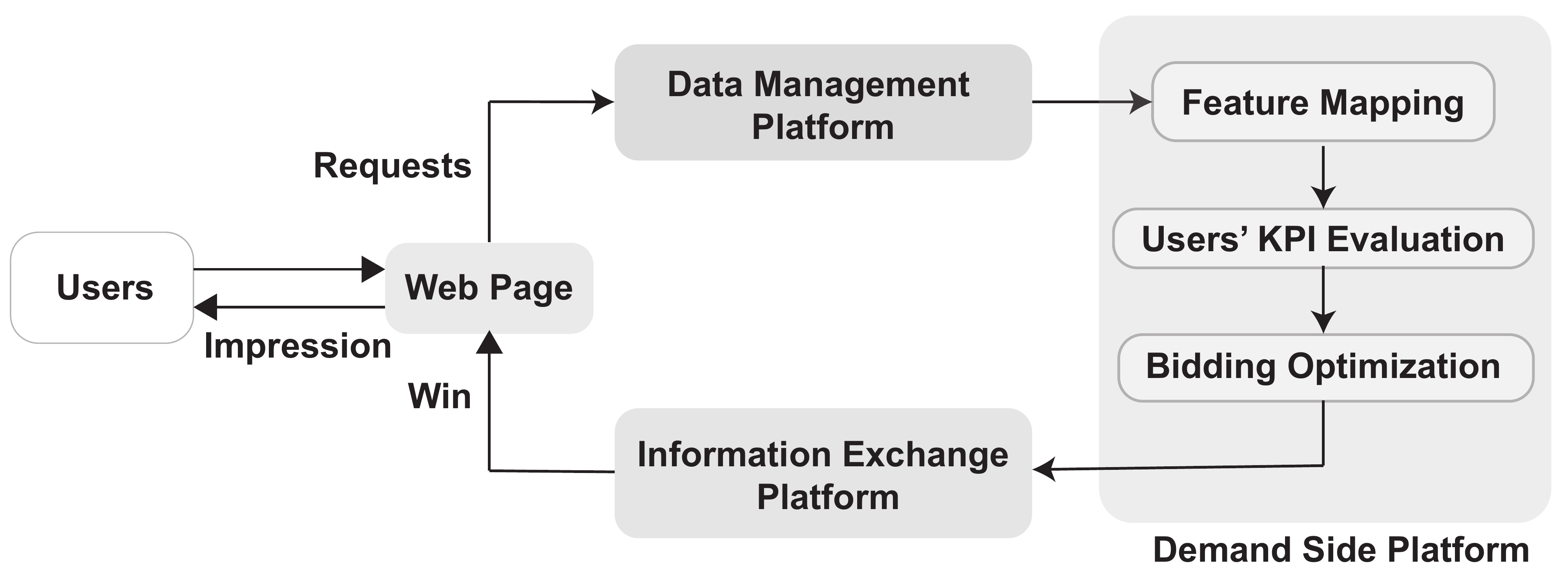}
    \caption{The Workflow of RTB. The web page detect the users' visits and send the ad requirement to the data management platform. DSP provide optimal bids based on user's and advertisement  historical data. The highest bidding price in the information exchange platform will win the impression and pay the second-highest price.}
    \label{Workflow}
\end{figure}

Advertising media has transitioned from television to online apps as the Internet has grown in popularity, making real-time advertising more accessible and prevalent. 
Advertisers have realized that the Internet allows them more flexibility and control over their advertising materials. Therefore it has become a widely used advertising medium. According to the advertising revenue report\cite{xu2021payment}, programmatic advertising (automated buying and selling of online advertising) takes up about 81\% of \$57 billion non-search advertising in 2019. \\
\indent Real-Time Bidding(RTB) based advertisement auction plays an essential role in daily life and becomes the new paradigm for internet advertising\cite{muthukrishnan2009ad}. Advertisers and media buying agencies use DSP to place bids, which allows them to bid automatically. 
For each request from the users, the advertiser in DSP must give the bid price within 100 milliseconds\cite{wang2016display}. DSP can have the information of the coming request, including the position of the advertisement, the information of the users, etc. DSP enable users to optimize based on pre-defined key performance indicators(KPIs) such as cost per click (CPC), cost per action (CPA), and their budget. 
  \\
\indent New challenges have arisen as a result of the introduction of RTB and header bidding technologies. Typically, there are two types of dynamic auctions, first-price auction, and second-price auction. In a first-price auction, multiple advertisers compete for impressions, with the highest bidder winning. Second-price auctions were created to allow marketers to bid up to the maximum amount allowed by their budget. Exploring the optimal bid price in a first-price auction is tricky as the advertiser will know only their bidding price and whether they win or not.
However, in a second-price auction, the second-highest price will be charged for the winner, which means the winner could know two information: \textbf{(1)} its bid price and \textbf{(2)} the second-highest bid price from the market. The bidding becomes more competitive in a second-price auction because the DSP has a greater grasp of the situation by access to additional data. The workflow of RTB is described in the Fig \ref{Workflow}.

In RTB campaigns, predicting the optimal bid price for each impression is one of the most common challenges. The bid request is won by the advertiser, who submits the highest bid price.
Advertisers’ objective is 
 to win as many impressions as possible
under the budget constraint by running the customized bidding strategies on DSP.

\indent 
The pacing model is a prominent sort of bidding strategy for branding campaigns that focuses on budget control \cite{xu2015smart, lee2013real, agarwal2014budget}. The game theory approach has been widely used in RTB problems, especially in the second price auction. Any pure strategy\cite{purestrategy,maskin1999nash} will be detected by others so that advertisers can modify their bid strategy and win effectively. Therefore, researchers often come up with mixed strategy\cite{mixed} like operational research or parameter optimization problem\cite{zhang2016optimal}. In addition, the efforts in \cite{1,21,32,33} attempt to use even model-free reinforcement strategy to achieve higher performance. To deal with high dimension features, the advertisers apply the factorization method, such as DeepFM, IFM, FEFM, FWFM etc\cite{guo2017deepfm,yu2019ifm,pande2020fefm,FWFM}, which maps the high dimensional sparse data into latent space so that advertisers can learn users' behavior.  
\\
\indent In this paper, we proposed a multi-agent reinforcement learning architecture, combining functional optimization and a model-free algorithm. We give agents different states, rewards, and actions associated with winning probability, click-through rate, etc. The main contributions of this paper are:
\begin{enumerate}
    \item Designed functional optimization agents (FOAs) using Lagrange multiplier 
    \item Combination of deep reinforcement learning and functional optimization to learn the agent's causality.
    \item Demonstrate the effectiveness of the functional optimization using the multi-agent bidding scenario.
\end{enumerate}
The rest of this paper is organized as follows: we introduce some related works and assumptions in the next section, followed by functional optimization in Section III. We present the multi-agent reinforcement learning architecture and the detailed design of the FOA in Section IV. We evaluate the FOA in simulated data and analyze the experiments result in Section V. Finally, we conclude our work and future works in Section VI.

\section{Related Work}
\label{related_work}

\subsection{Key Performance Indicator}
KPI can be click-through rate(CTR) or click conversion rate(CVR) \cite{32}, and it influences bidders in making bidding decisions. Advertisers are more likely to bid at higher prices for users with higher KPIs to win these valuable impression opportunities. As KPI estimation is the fundamental step of RTB, its accuracy directly affects the estimation of the bidding price. Traditional KPI prediction models try to use machine learning methods, such as regression, random forest, and gradient boosting\cite{randomforest, boosting}. However, these methods can only learn the users' features on the surface and cannot explain the high order feature interactions. Therefore, Rendle, in 2010, came up with the factorization machine, which can map the original features into high dimension latent vectors and determine their relationships in second or higher levels \cite{rendle2010factorization}. After that, the deep learning method increased rapidly and was gradually replaced. In 2017, Guo et al. proposed the DeepFM with integrating the Factorial Machine and deep learning and achieve good performance\cite{guo2017deepfm}. Since that, the modified method boosted\cite{pande2020fefm,yu2019ifm}.

\subsection{Winning Probability}
The winning probability is one of the keys to finding the optimal price solutions. It works together with KPIs to influence bidding decisions. Computing the auction winning probability seems simple because the cumulative sum of winning densities is the winning probability of the corresponding price. And this cumulative density distribution is always monotonically increasing; However, it is biased because the distribution is only learned from winning data. On the contrary, Zhang et al., in \cite{(1)2} models the auction winning probability with a bid landscape based on nonparametric survival analysis for second-price auction. It learns from the winning bids as well as from the losing bids so that the learned winning probability distribution is unbiased. 

\subsection{Operational Approachs}
The optimal RTB is typically treated as a dynamic programming problem. Within the constrain of the budget, DSP attempts to win more impressions with less cost, i.e., reach-based campaign. Therefore, \cite{OR1,OR2,OR3,OR4,OR5} design the objective function to maximize the expected impression value and minimize the expected cost. Many authors often utilize the Lagrange multiplier method to solve this problem by converting it to a dual problem. However, they always assume the fixed distribution for some terms in the equation. For instance, Zhang et al., in \cite{21} assume a convex relationship of wining probability and bid price. Karlsson et al., in\cite{adInventory} propose a concave relationship of impression value and optimal bid price.

\subsection{Model-Free Approachs}
 To optimize the RTB campaign, Several criteria are examined, including the duration of the campaign, the preferences of each advertising group, the rivals' bids and tactics, the publishers' reserve pricing, and the number of networks \cite{6960761}. Therefore, optimal RTB is a mixed-strategy game. There is no pure strategy to bid. With the fast development of reinforcement learning, \cite{RL1} has tried the model-free method and shown excellent performance. Alibaba Group also proposes a multi-agent reinforcement learning\cite{jin2018real}. In the model-free methods, advertisers are considered agents and the state's budget consumption rate, etc. Each advertiser can design its reward function to maximize the profit. Therefore, for a given request, the agent will inference the bid price. After the information is exchanged in the campaign, the agent will get the feedback and calculate the reward, then update the action network and repeat. Based on previous studies, reinforcement learning achieves excellent performance and is distributed on most platforms. However, the model-free approach is closed to the statistical analysis. Fu et al., in \cite{NIPS_b} proved that unconstrained bidding strategies are hard to learn. We would better focus on even monotone bidding strategies. To solve this problem, we propose a functional optimization reinforcement learning method. 
\subsection{Assumptions and Notations}
\begin{table}[!t]
\caption{\textbf{Symbols and Descriptions}}
\centering
\vspace{0.4mm}
\setlength{\tabcolsep}{5mm}{
\begin{tabular}{llp{4.5cm}}
\hline
\vspace{0.4mm}
 & Symbols & Description                              \\ \hline
 
 & $\boldsymbol{x}$     & One bid request represented by its features  \\
 & $P_\mathbb{X}(\boldsymbol{x})$    & Probability density function of bid request $\boldsymbol{x}$     \\
 & $c(\boldsymbol{x})$     & The predicted key performance indicator(KPI) for advertisers if winning the bids. It could be CTR or CVR $\boldsymbol{x}$    \\
 & $P_\mathbb{C}(c(\boldsymbol{x}))$      & Probability density function of $c$  \\
 & $B$     & Total Budget of one campaign \\ 
 & $N$      & Number of requests in one campaign  \\
 & $b(c(\boldsymbol{x}))$     & Bid price function which determines the                                    bidding strategy by taking $\boldsymbol{x}$ and                                $c(\boldsymbol{x})$ as input. Assume the following dependency: $\boldsymbol{x}$$\rightarrow$$c$$\rightarrow$$b$ \\
 & $w(b(c(\boldsymbol{x})))$      & Win probability estimation function. Calculate from the dependency assumption:  $\boldsymbol{x}$$\rightarrow$$c$$\rightarrow$$b$$\rightarrow$$w$
                        \\
\hline
\end{tabular}}
\end{table}

\indent To make the equations in the next section solvable, we have made the following assumptions:
\begin{itemize}
    \item Assume KPI estimation $c(\boldsymbol{x})$ is determined by features $\boldsymbol{x}$. 
    \item Assume the bidding strategy only depends on KPI estimation $c(\boldsymbol{x})$, like the previous work in \cite{21}.
    \item Assume the winning probability function with respect to bidding price is monotonically increasing and only influenced by the bidding price. Previous works \cite{Joint_optimization,32,Keyword_Auctions} also make the the similar assumption. 
\end{itemize}
above assumptions makes feature $\boldsymbol{x}$ influence the win probability with the following process: $\boldsymbol{x}$$\rightarrow$$c$$\rightarrow$$b$$\rightarrow$$w$.
\section{Functional Optimization}
In this section, we formulated the RTB problem mathematically\cite{32}. 
For a given bid request with the high dimensional feature vector $\boldsymbol{x}$, DSP's bidding engine decides to participate in the auction and how much to pay for the current request to win. This paper aims to provide a combined optimization framework that considers user reaction, market competitiveness, and bidding strategy to maximize the advertiser's total profit.
\\

\noindent\textbf{Optimal Bidding Function} The optimal bidding function $b_{optimal}()$ is  necessary to maximize the estimated impression value per campaign while staying within the campaign's budget $B$.
Therefore, the RTB problem can be formulated as an optimization problem, \cite{32} as follows:
\begin{align}
\label{problem_formulation}
\boldsymbol{b}_{optimal} &=\underset{b(c(\boldsymbol{x}))}{\arg \max}\;\; N\int_{\boldsymbol{x}} c(\boldsymbol{x}) w(b(c(\boldsymbol{x})))P_\mathbb{X}(\boldsymbol{x}) \boldsymbol{d} \boldsymbol{x}  \\
& \text {s.t. } N\int_{\boldsymbol{x}} b(c(\boldsymbol{x})) w(b(c(\boldsymbol{x}))) P_\mathbb{X}(\boldsymbol{x}) \boldsymbol{d} \boldsymbol{x} \leq B \nonumber
\end{align}

 The product of the KPI $c(\boldsymbol{x})$ and winning probability $w(b(c(\boldsymbol{x})))$, gives the expected KPI per impression auction. $P_\mathbb{C}(c(\boldsymbol{x}))$ is the prior probability distribution of feature vectors $\boldsymbol{x}$. Now, the objective is to marginalize winning probability and KPI over feature space and then multiply by the total number of requests to yield the expected impression value per campaign.

\indent According to Zhang, et al \cite{32} and our assumptions, the above optimization problem can be expressed with respect to KPI prediction $c(\boldsymbol{x})$ or $c$:
\begin{align}
\label{KPI_based}
\boldsymbol{b}_{optimal}&=\underset{b(c)}{\arg \max}\;\;N\int_{\boldsymbol{c}} c w(b(c)
P_\mathbb{C}(c)\boldsymbol{d} \boldsymbol{c}  \\
& \text {s.t. } N \int_{\boldsymbol{c}}b(c) w(b(c))
P_\mathbb{C}(c)\boldsymbol{d} \boldsymbol{c}\leq B \nonumber
\end{align}
the Lagrangian function of equation (\ref{KPI_based}) is
\begin{align}
\label{Lagrangian_function}
\mathcal{L}(b(c), \lambda)= \int_{c} &c w(b(c)) P_\mathbb{C}(c) \boldsymbol{d} \boldsymbol{c} \\ \nonumber
&- \lambda \int_{c} b(c) w(b(c)) P_\mathbb{C}(c) \boldsymbol{d} \boldsymbol{c}+\frac{\lambda B}{N}  
\end{align}
where $\lambda$ is the Lagrange multiplier. After solving the Euler-Lagrange function (\ref{Lagrangian_function}),
\begin{align}
    c P_\mathbb{C}(c) \frac{\partial w(b(c))}{\partial b(c)}-\lambda P_\mathbb{C}(c)\left[w(b(c))+b(c) \frac{\partial w(b(c))}{\partial b(c)}\right]=0 
\end{align}
\begin{align}
    \lambda w(b(c))=[c-\lambda b(c)] \frac{\partial w(b(c))}{\partial b(c)}
    \label{simplified_lagrange_multiplier}
\end{align}
\indent As proposed by Zhang et. al \cite{32}, the winning probability function, which is monotonically increasing and deemed to have a concave shape is assumed to have the form: $w(b(c)) = \frac{b(c)}{a+b(c)}$. However, only one parameter $a$ gives agents limited freedom to learn the $\lambda$ from the states or take $\lambda$ as a reward for evaluating its bidding quality. The agents will have low expressiveness and performance. To keep the concave shape of the winning probability function while providing sufficient parameters and making the function is easily differentiable. Instead, we assume the winning probability function has the standard cubic form:
\begin{align}
\label{cubic_curve}
    w(b(c))) = m_{1}b^3+m_{2}b^2+m_{3}b+m_{4}
\end{align}

\indent Take the partial derivative of $w(b(c)))$ with respect to $b$ and substitute into equation (\ref{simplified_lagrange_multiplier})
\begin{align}
    \lambda(m_{1}b^3+m_{2}b^2+m_{3}b+m_{4}) = (c-\lambda b)(3m_{1}b^2+2m_{2}b+m_{3})
\label{7}
\end{align}
where $b$ is the bidding price determined by the estimated KPI $c$. Rearranging the above equation gives us the Lagrange multiplier
\begin{align}
    \lambda = \frac{c(3m_{1}b^2+2m_{2}b+m_{3})}{4m_{1}b^3+3m_{2}
    b^2+2m_{3}b+m_{4}}
    \label{Lagrange_multiplier}
\end{align}
\section{Multi-agent Reinforcement Learning Architecture}
\begin{figure}[!t]
     \centering
    \includegraphics[width = 3.5in]{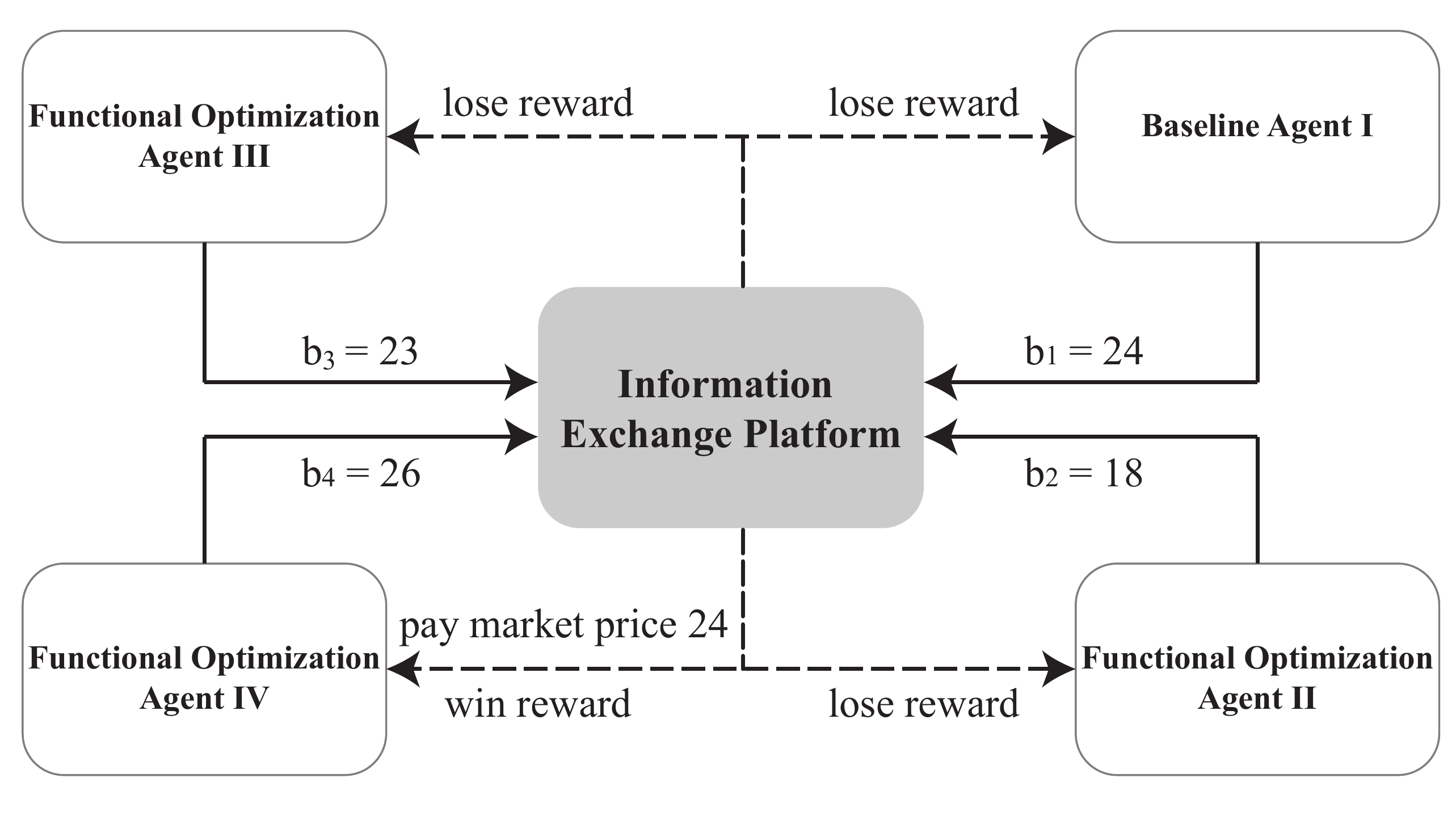}
     \caption{The MARL Architecture for Second Price Auction. For each request in the campaign, agents will give an optimal bid price based on their strategy. The highest bid price wins the impression on the information exchange platform. Finally, the agents get the feedback from the results and do the strategy update.}
    \label{MARL}
\end{figure}

To find the optimal solution of the bidding function, we designed a bidding strategy by combining a model-free reinforcement learning model with resultant $\lambda$ as a tuning parameter for the bidding function. Specifically, we utilized the Deep-Q learning model to determine the optimum action for each state. Fig \ref{MARL} shows the proposed multi-agent reinforcement learning (MARL) architecture. The following are the steps of the proposed model in more detail:
\subsection{Environment Setup}
Due to the stochastic bidding environment of each campaign, it is difficult to evaluate the performance of advertisers and verify the efficacy of the operating method in the real bidding market. As a result, we depict the market with a virtual environment. \newline
We designed four agents in the bidding environment: three functional optimization agents (FOAs) and one baseline agent. The winning agent will have two information: its bid price and the second-highest bid price from the market. The other agents can only know they are losing this request with their bidding prices.
\subsection{Agents Setup}

The functional optimization agents are designed as given in Algorithm 1, namely, the Lagrange multiplier as shown in equation (\ref{Lagrange_multiplier}).  For functional optimization agents, the first agent consider the Lagrange multiplier as a component of state, which is updated over each request. The second agent consider Lagrange multiplier as its action, aiming to learn the potential best Lagrange multiplier during the bidding so that its actions will maximize the objective function (\ref{KPI_based}). The third agent treats the Lagrange multiplier as one component of its reward function, reflecting the previous bidding quality. The winner has to pay the second-highest bid price for current request. \\
\indent However, when considering the time complexity, it is impractical to numerically fit the cubic winning probability curve (\ref{cubic_curve}) after per action for \textbf{FOA II} and \textbf{IV}. Additionally, the pre-experiments also demonstrate the fitted continuous curves sometimes are not monotonically increasing. To address these problems, we discretize the equation (\ref{simplified_lagrange_multiplier}) by substituting a partial derivative $\frac{\partial w(b(c))}{\partial b(c)}$ with 
$$\frac{\Delta w(b)}{h},\;where\; \Delta w(b) = w(b_{i}) - w(b_{i-1})$$
$b_{i}$ and $b_{i-1}$ are the sequential bid price in the current bid set and $h$ is their difference. So for \textbf{FOA III} which aims to calculate bidding price based on action $\lambda$, we have
\begin{equation}
    b = \frac{c\Delta w(b) - h\lambda w(b)}{\lambda \Delta w(b)} 
\label{9}
\end{equation}
and for \textbf{FOA II} and \textbf{IV} which take $\lambda$ as reward, the equation (\ref{simplified_lagrange_multiplier}) becomes:
\begin{equation}
    \lambda = \frac{c\Delta w(b)}{hw(b)+b\Delta w(b)} 
\label{10}
\end{equation}
where $w(b)$, can be biased or unbiased, is discrete probability distribution learned simultaneously over the campaigns.
\subsection{Deep Q-learning}
Deep Q-learning uses neural network which
maps input states to (action, Q-value) pairs. The proposed DQN setup is as follows: \\

\begin{itemize}
    \item $State:=$ [Remaining Budget,Budget Consumption Rate, Win Rate, Step].
    \begin{itemize}
        \item $R_B$ denotes remaining budget 
        \item The budget consumption rate, $\frac{\sum win \;bid}{200}$, is updated after per bids. (Budget consumption rate is calculated in a deque of which the capacity is 200)
        \item Win rate represents the overall win rate in one campaign
        \item Step represents the progress of one campaign. The maximum step for each agent is the number of total bids $N$.
    \end{itemize}
    
    \item \textit{Action}. Each agent will choose an bid price as the output except the \textit{FOA III}. The action of \textit{FOA III} is to leverage $\lambda$. The bid price of this agent will be numerically approximated from equation (\ref{9}). 
 
    \item \textit{Reward}. If the agent wins this request, it will be rewarded as 5. Otherwise, it will be rewarded as -1. If the agent wins most at the end of each campaign, it will be rewarded as 200. For the \textbf{FOA IV}, the reward function will contain an additional $\lambda$ which is directly obtained from equation (\ref{Lagrange_multiplier}).
    
    \item \textit{Action-value} $f:Q(s,a)$. We utilize the Deep Q-Network as each agent's action-value function, which has three hidden layers with 128 units for each. After every request, the information will be sent to the experience replay buffer. The capacity of the replay buffer is 1000. After every 20 bids, we will randomly sample 20 bidding information to learn. To train this network, firstly, we will calculate the current expected value for the action. 
    \begin{align}
        Q^{*}(s, a)=\mathbb{E}_{s^{\prime} \sim \mathcal{E}}\left[r+\gamma \max _{a^{\prime}} Q^{*}\left(s^{\prime}, a^{\prime}\right) \mid s, a\right]
    \end{align}
    where $\gamma$ is the importance of future, $s^{\prime}$ and $a^{\prime}$ represent the next state and action, $r$ indicates the reward. So the loss function becomes
    \begin{align}
        y_{j}=\left\{\begin{array}{l}
r_{j} \text { for terminal } \mathrm{s}_{j+1} \\
r_{j}+\gamma \max _{a^{\prime}} Q\left(\mathrm{~s}_{j+1}, \mathrm{a}^{\prime} ; w\right) \text { for non-terminal } \mathrm{s}_{j+1}
\end{array}\right.
    \end{align}
    $w$ represents the weight of the Q-Network. To update the network, we have to find the gradient of this loss function.
    \begin{align}
        \nabla_{w_{j}} L_{j}(w_{j})= &\mathbb{E}_{s, a \sim \rho(\cdot);s^{\prime} \sim \mathcal{E}} \Big[(r+\gamma \max _{a^{\prime}} Q(s^{\prime}, a^{\prime} ; w_{j-1}) \nonumber \\ 
        &-Q(s, a ; w_{j})) \nabla_{w_{j}} Q(s, a ; w_{j})\Big] 
    \end{align}
\end{itemize}

\begin{algorithm}[H]
\caption{Baseline and Functional Optimization Agents}
\label{algorithm1}
\begin{algorithmic}
\State  $N_c$ := number of campaigns
\State  Initialize experience replay D with capacity $\frac{N}{5}$ 
\State  Initialize action-value Q-network
\State  \textbf{Input} Operation research function, Bidding environment
\State  \textbf{Output} Q-network $\boldsymbol{Q}(s,a,w)$
  \For {$i = 1: N_c$}
\State  Agents setup 
    \For{$j = 1: N$}
\State Select random action $a_{j}$ with probability $\epsilon$ 
\State  Otherwise select action $a_{j} = \underset{a}{max}Q(s_{j},a;w)$ 
\If{\textbf{Baseline I}}
\State {$b := a_{j}$
\State  store bidding results into varuable \textit{result}
\State calculate the reward function r}
        \ElsIf{\textbf{FOA II:}}
\State         {$b := a_{j}$
\State         store bidding results into variable $result$
\State         update $w(b)$ using $result$
\State         update $\lambda$ from equation (\ref{10}) using $c,w(b),h$
\State         calculate the reward function $r$
\State      update state using $\lambda$}
   \ElsIf{\textbf{FOA III:}} 
\State  $\lambda := a_{j}$ 
\State        solve the equation (\ref{9}) to get the bid price $b$ 
\State store bidding results into variable \textit{result}
\State        update $w(b)$ using $result$
\State         calculate the reward function $r$
         
    \ElsIf{\textbf{FOA IV:}}
\State        $b := a_{j}$
\State         store bidding results into variable $result$
\State         update $w(b)$ using $result$
\State         update $\lambda$ from equation (\ref{10}) using $c,w(b),h$
\State        calculate the reward function $r$ using $\lambda$

\EndIf
\State        Store transition($s_{j},a_{j},r,s_{j+1}$) in D
\State        Set state $s_{j+1} = s_{j}$
\State        Sample random minibatch from D
\State        Set 
\begin{align}
    y_{j}=\left\{\begin{array}{l}
r_{j} \text { terminal } \\
r_{j}+\gamma \max _{a^{\prime}} Q\left(\mathbf{s}_{j+1}, \mathbf{a}^{\prime} ; w\right) \text { non-terminal }
\end{array}\right.
\end{align}
           
\State  Perform a gradient descent $\left(y_{j}-Q\left(s_{j}, a_{j}; w\right)\right)^{2}$
\EndFor
\EndFor
\end{algorithmic}
\end{algorithm}

\section{Experiments and Results}
\subsection{Experiments Setup}
To demonstrate the efficiency of the proposed framework, we simulate ten  campaigns (second-price autions) sequentially ($N_c=10$) and assume that  agents bid together.
\\
\indent As discussed in Section \ref{related_work}, both biased and unbiased winning probability approaches are used and two sets of experiments are conducted separately. In addition, we set diverse budgets to imitate the real-world situation of budget shortage and abundance to test the durability of the FOAs in varied settings. For campaigns with biased winning probability, we set the number of bid requests $N_{biased} = 50K$ and assign budget $B_{baised}=\{250K,500K,750K,100K,1250K\}$ in dollar for each campaigns. While for campaigns utilizing unbiased winning probability, due to the time complexity in camparing global prices and computing unbiased probability, we set $N_{unbiased}=1K$ and $B_{unbiased}=\{5K,10K,15K,20K,25K\}$ so that $$
\frac{B_{biased}}{N_{biased}} = \frac{B_{unbiased}}{N_{unbiased}}
$$
The bidding price interval is discretized from \$10 to the cap price \$100, where the minimum difference is \$1.  All experiments are performed on a Inter$\circledR$ i9-10850K 10-core CPU with 16 GB RAM.

\subsection{Parameter Setup}
We set $\epsilon$ taking about 10K requests to decay from 1 to 0.01. $\gamma$ = 0.95, batch size = 20. For equation (\ref{9}) (\ref{10}), $h=1$ for biased winning probability where $h$ is dynamic for unbiased probability. For CTR $c=0.001$ which is a constant after checking the real bidding data.  
\subsection{Results}
\begin{table}[t]
\centering
\caption{Average Win Rate and Surplus Over 10 Campaigns with Biased Winning Probability}
\begin{tabular}{r|llll}
\hline
\textbf{Budget}                         & \textbf{BA I} & \textbf{FOA II}  & \textbf{FOA III}          & \textbf{FOA IV}  \\  \hline
\multirow{2}{*}{\$250K}   & 12.90\%                   & 12.78\%          & \textbf{62.10\%}          & 12.22\%          \\
                           & \$23.59                   & \$22.51          & \textbf{\$48.19} & \$22.58          \\ \hline
\multirow{2}{*}{\$500K}   & 17.65\%                   & 18.08\%          & \textbf{46.73\%}          & 17.54\%          \\
                           & \$23.07                   & \$22.21          & \textbf{\$46.68} & \$21.41          \\ \hline
\multirow{2}{*}{\$750K}   & 24.73\%                   & \textbf{26.59\%} & 23.53\%                   & 25.15\%          \\
                           & \$22.67                   & \$24.12          & \textbf{\$81.19} & \$21.45          \\ \hline
\multirow{2}{*}{\$1000K} & 30.33\%                   & 30.47\%          & 8.71\%                    & \textbf{30.50\%} \\
                           & \$23.76                   & \$22.40          & \textbf{\$33.55} & \$22.51          \\ \hline
\multirow{2}{*}{\$1250K} & \textbf{34.72\%}          & 31.97\%          & 6.39\%                    & 26.91\%          \\
                           & \textbf{\$23.75} & \$21.82          & \$17.64                   & \$20.59          \\ \hline
\end{tabular}
\label{biased_table}
\end{table}

\begin{table}[t]
\centering
\caption{Average Win Rate and Surplus Over 10 Campaigns with Unbiased Winning Probability}
\begin{tabular}{r|llll}
\hline
\multicolumn{1}{l|}{\textbf{Budget}} & \textbf{BA I} & \textbf{FOA II}  & \textbf{FOA III} & \textbf{FOA IV}  \\ \hline
\multirow{2}{*}{\$5K}   & 25.37\%                   & 20.26\%          & 22.90\%          & \textbf{31.47\%} \\
                      & \$22.16                   & \$20.05          & \textbf{\$30.91} & \$24.54          \\ \hline
\multirow{2}{*}{\$10K}  & 26.84\%                   & 25.15\%          & 13.78\%          & \textbf{34.23\%} \\
                      & \$23.05                   & \$25.07          & \$26.96          & \textbf{\$30.13} \\ \hline
\multirow{2}{*}{\$15K}  & 22.36\%                   & 27.44\%          & 12.32\%          & \textbf{37.88\%} \\
                      & \$18.24                   & \$22.31          & \$26.85          & \textbf{\$31.68} \\ \hline
\multirow{2}{*}{\$20K}  & \textbf{31.27\%}          & 30.94\%          & 10.11\%          & 27.68\%          \\
                      & \textbf{\$25.54}          & \$16.98          & \$22.24          & \$20.00          \\ \hline
\multirow{2}{*}{\$25K}  & \textbf{32.87\%}          & 29.32\%          & 8.95\%           & 28.86\%          \\
                      & \$21.21                   & \textbf{\$23.47} & \$17.73          & \$21.46          \\ \hline
\end{tabular}
\label{unbiased_table}
\end{table}

\begin{figure*}[t]
\centering
\subfigure{
    \includegraphics[width=4.5cm]{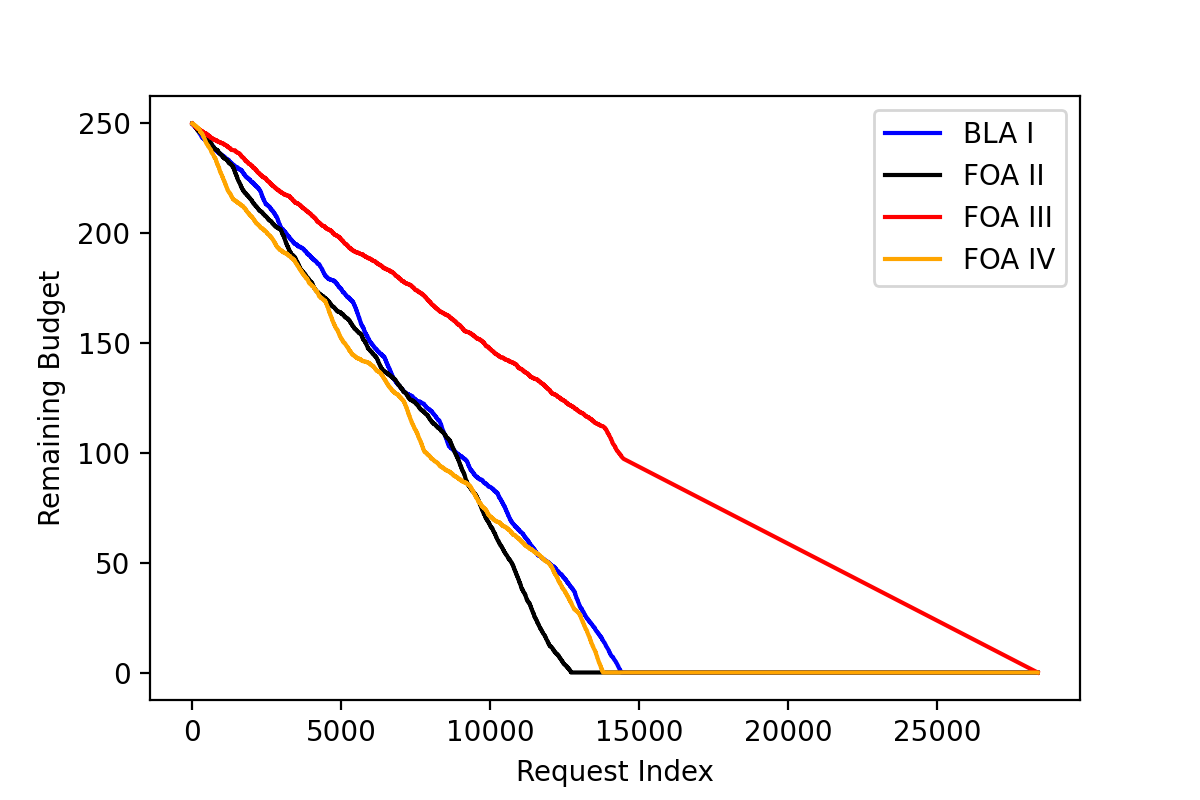}}\hspace*{-0.9em} 
\subfigure{    
    \includegraphics[width=4.5cm]{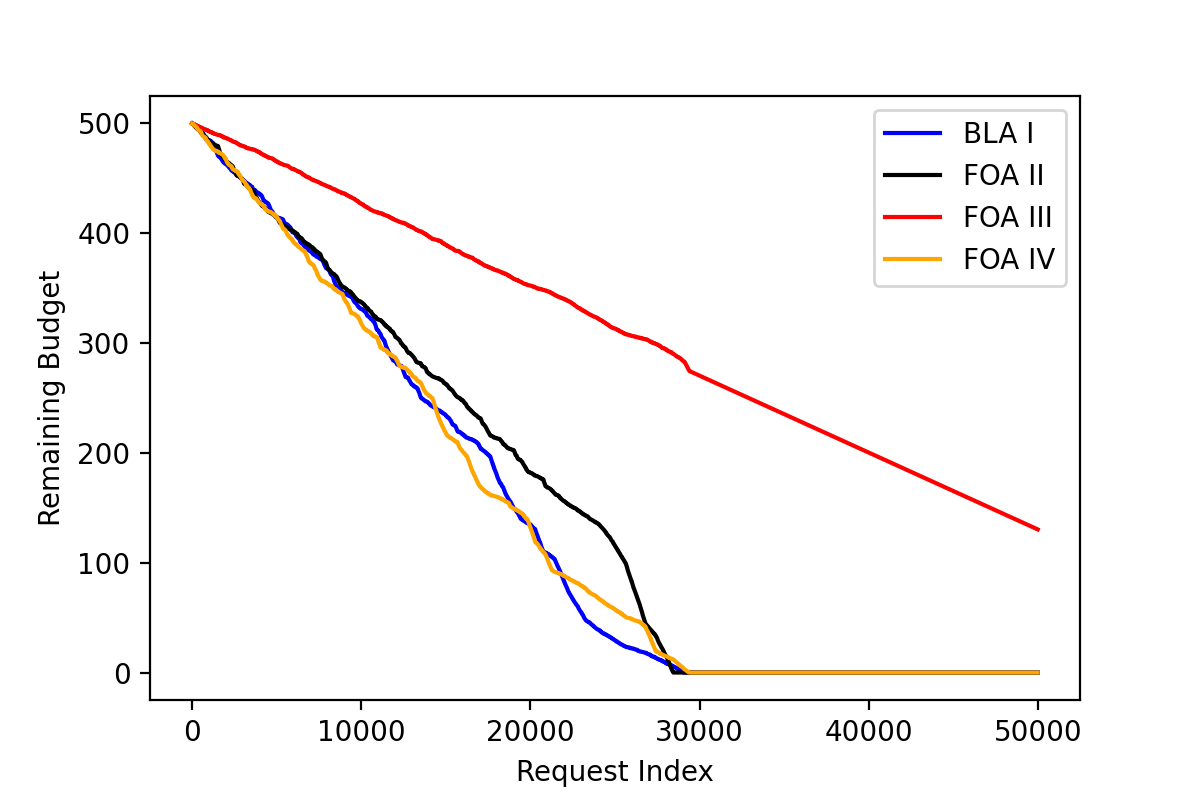}}\hspace*{-0.9em} 
\subfigure{
    \includegraphics[width=4.5cm]{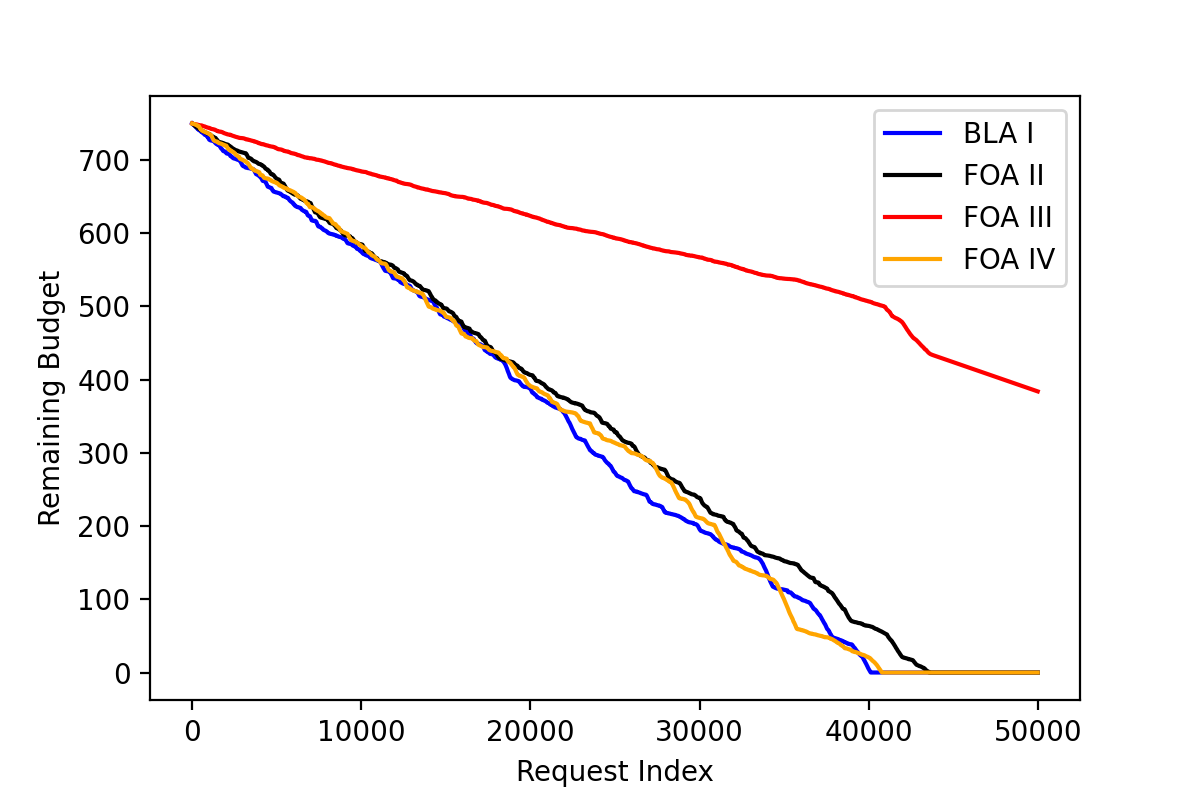}}\hspace*{-0.9em} 
\subfigure{
    \includegraphics[width=4.5cm]{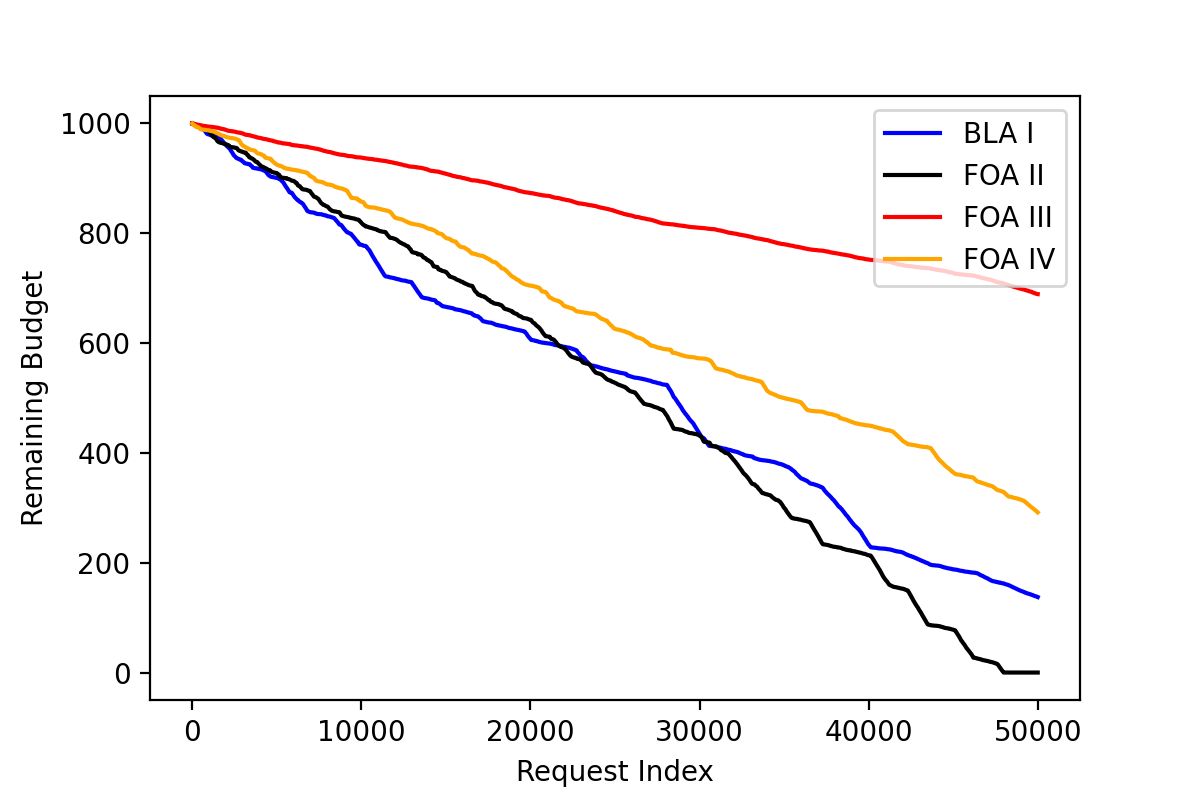}}\hspace*{-0.9em} 
    \\
    \vspace*{-0.9em}

\subfigure[(a) Budget=250K]{
    \includegraphics[width=4.5cm]{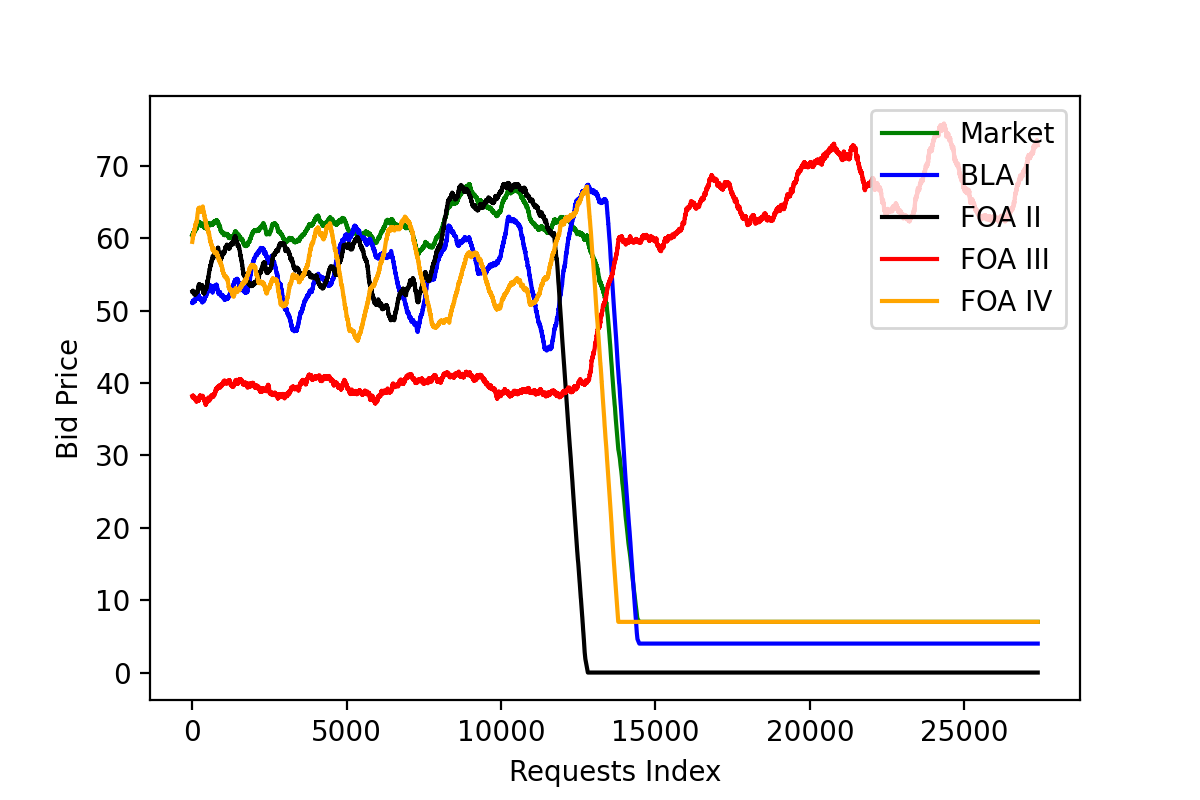}}\hspace*{-0.9em} 
\subfigure[(b) Budget=500K]{
    \includegraphics[width=4.5cm]{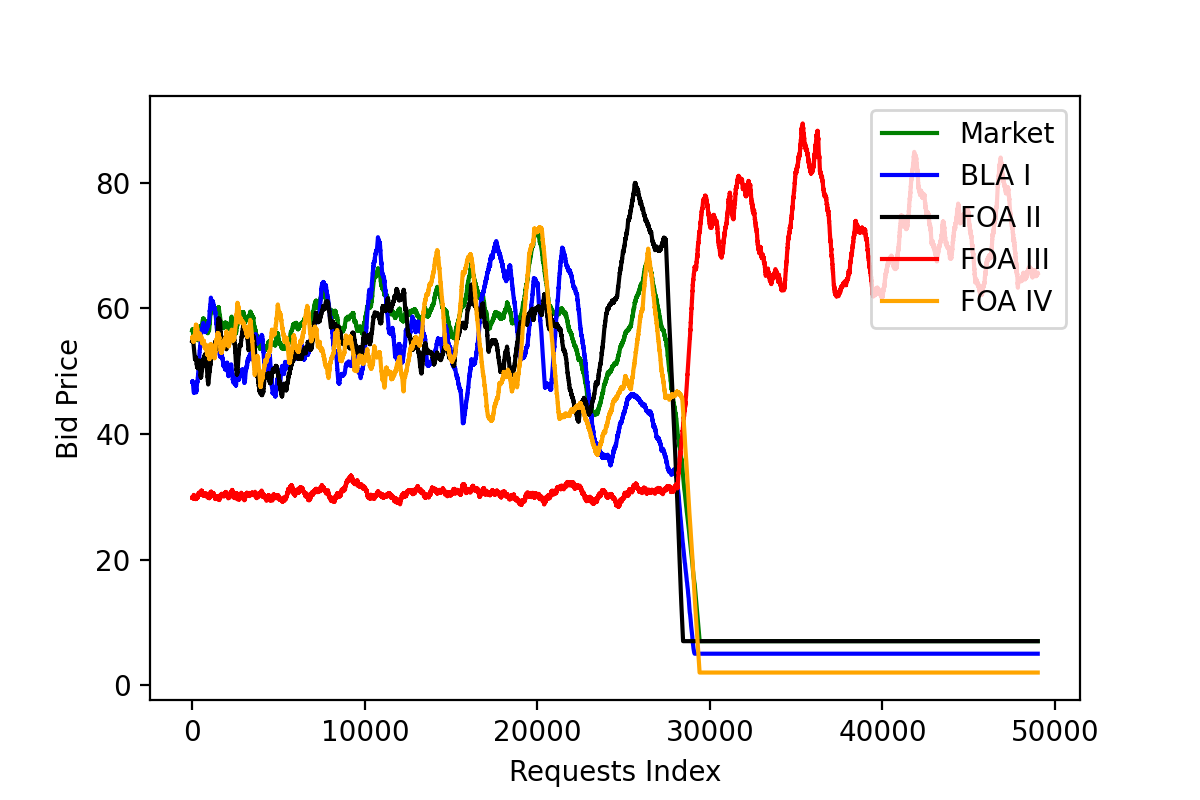}}\hspace*{-0.9em} 
\subfigure[(c) Budget=750K]{
    \includegraphics[width=4.5cm]{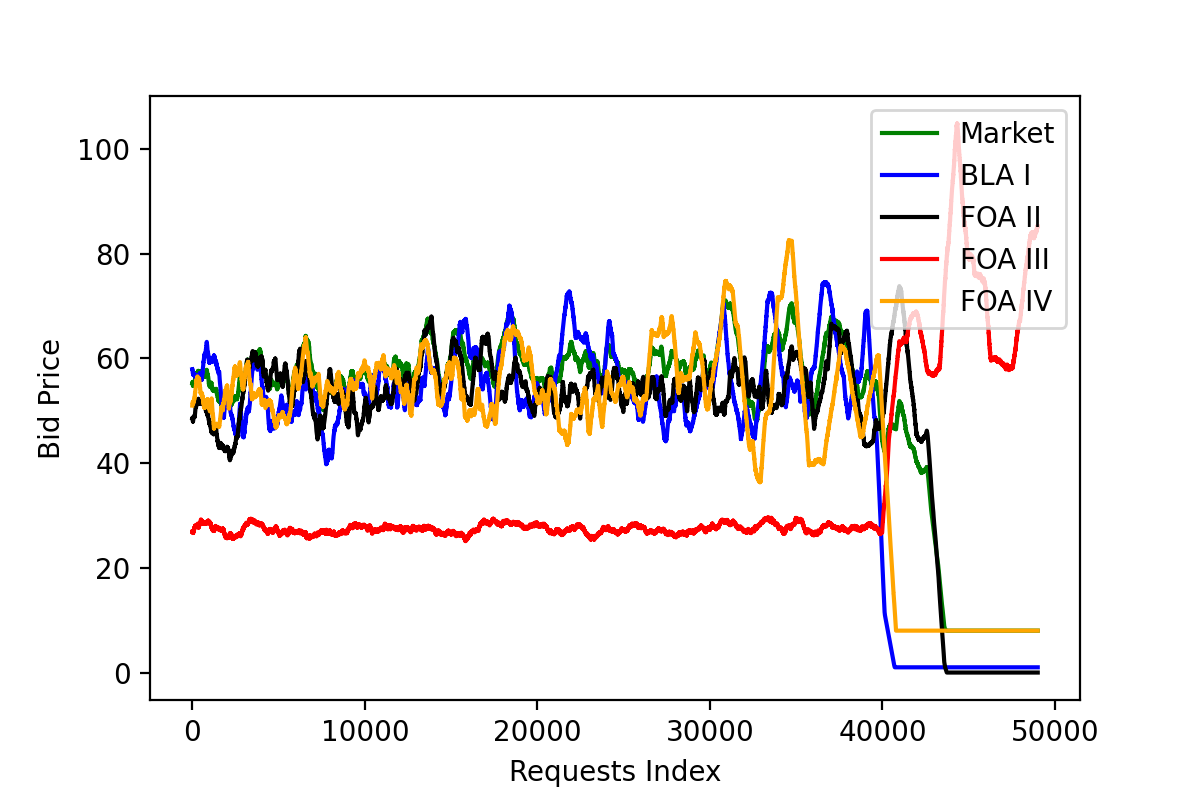}}\hspace*{-0.9em} 
\subfigure[ (d)Budget=1000K]{
    \includegraphics[width=4.5cm]{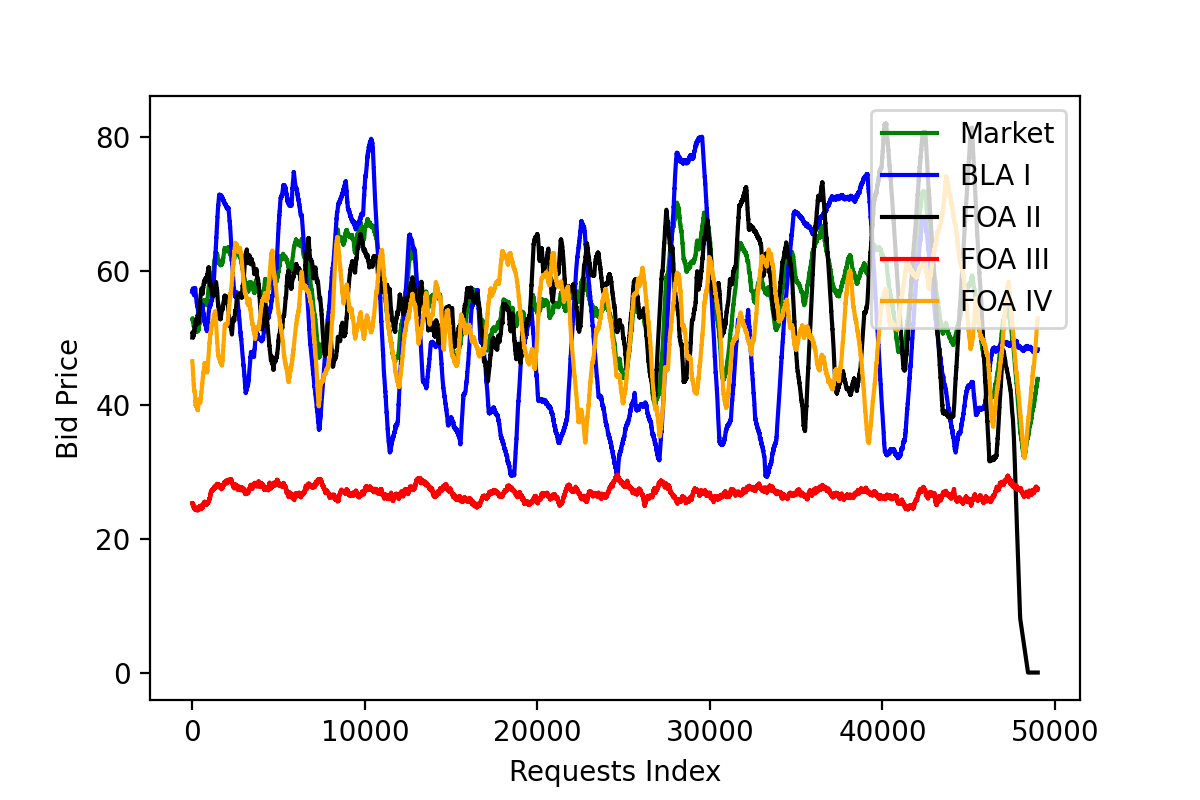}}\hspace*{-0.9em}
\caption{Biased algorithm comparison results: Budget Consumption (first row) and Distribution of the bid price and winning (second row).}
\label{results_biased}
\end{figure*}

\begin{figure*}[t]
\centering
\subfigure{    
    \includegraphics[width=4.5cm]{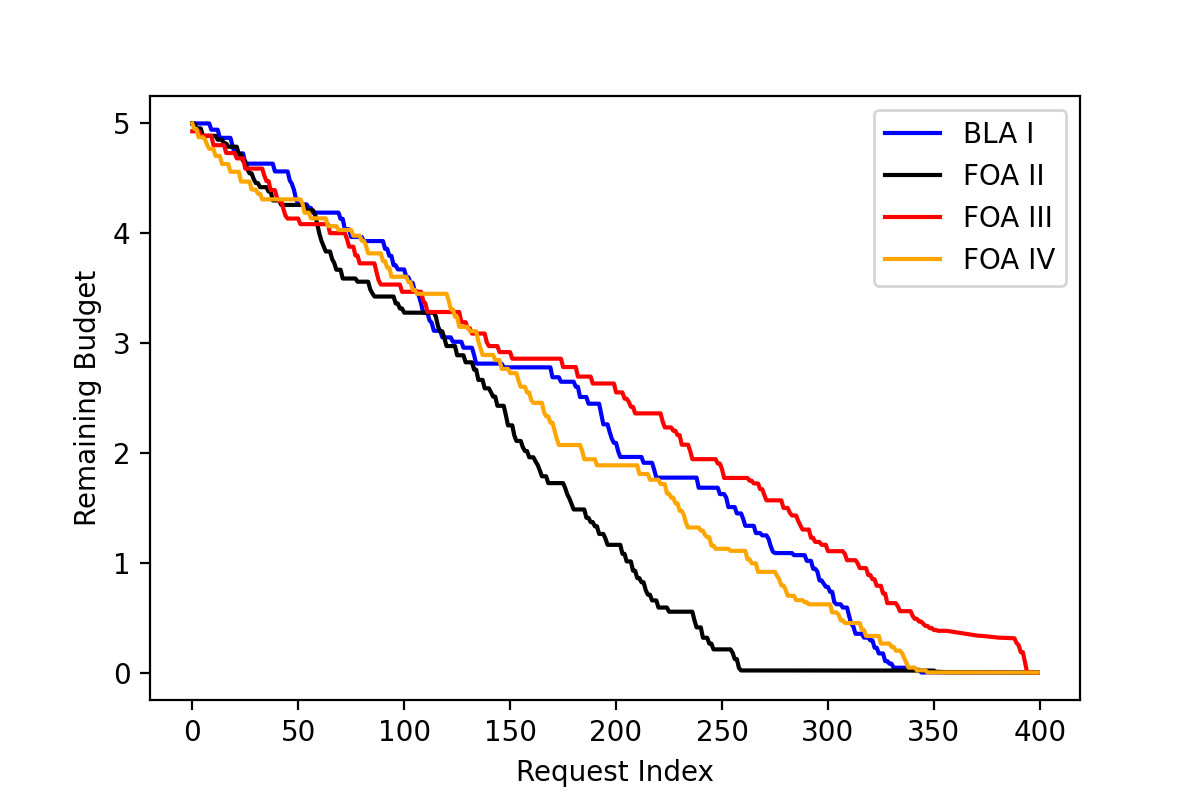}}\hspace*{-0.9em} 
\subfigure{
    \includegraphics[width=4.5cm]{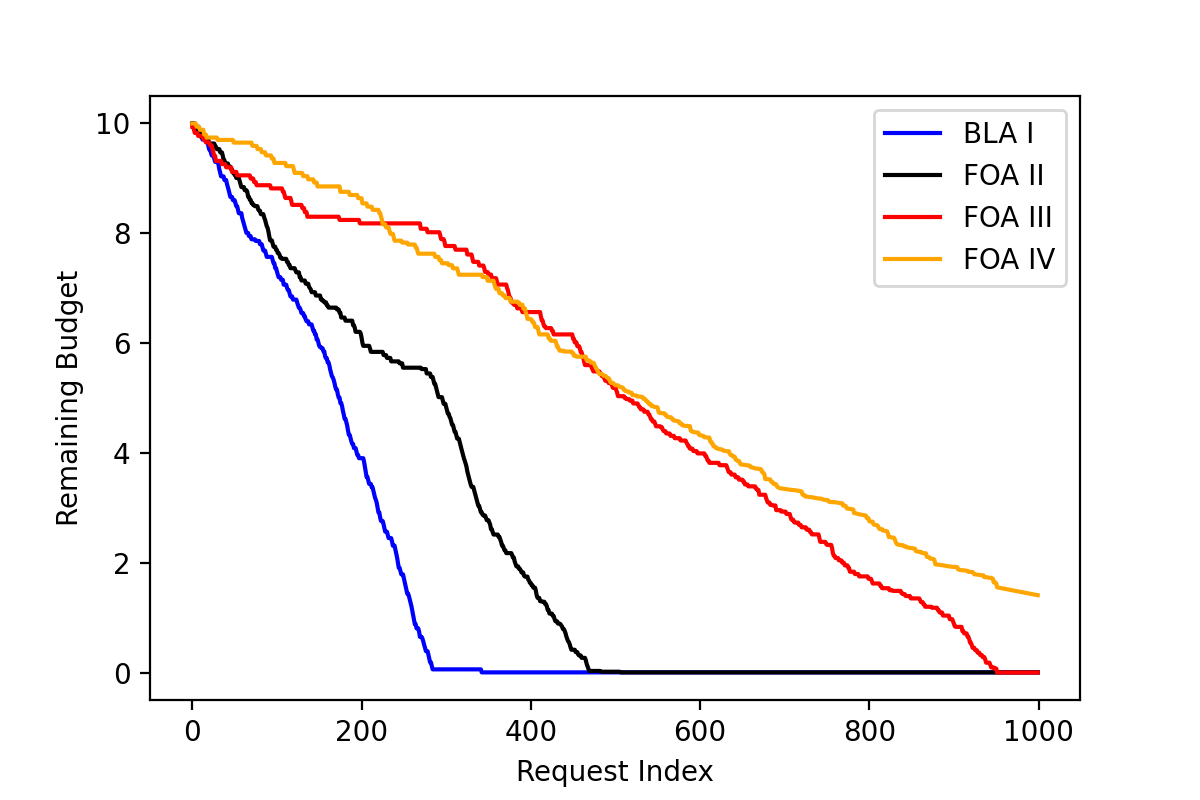}}\hspace*{-0.9em} 
\subfigure{
    \includegraphics[width=4.5cm]{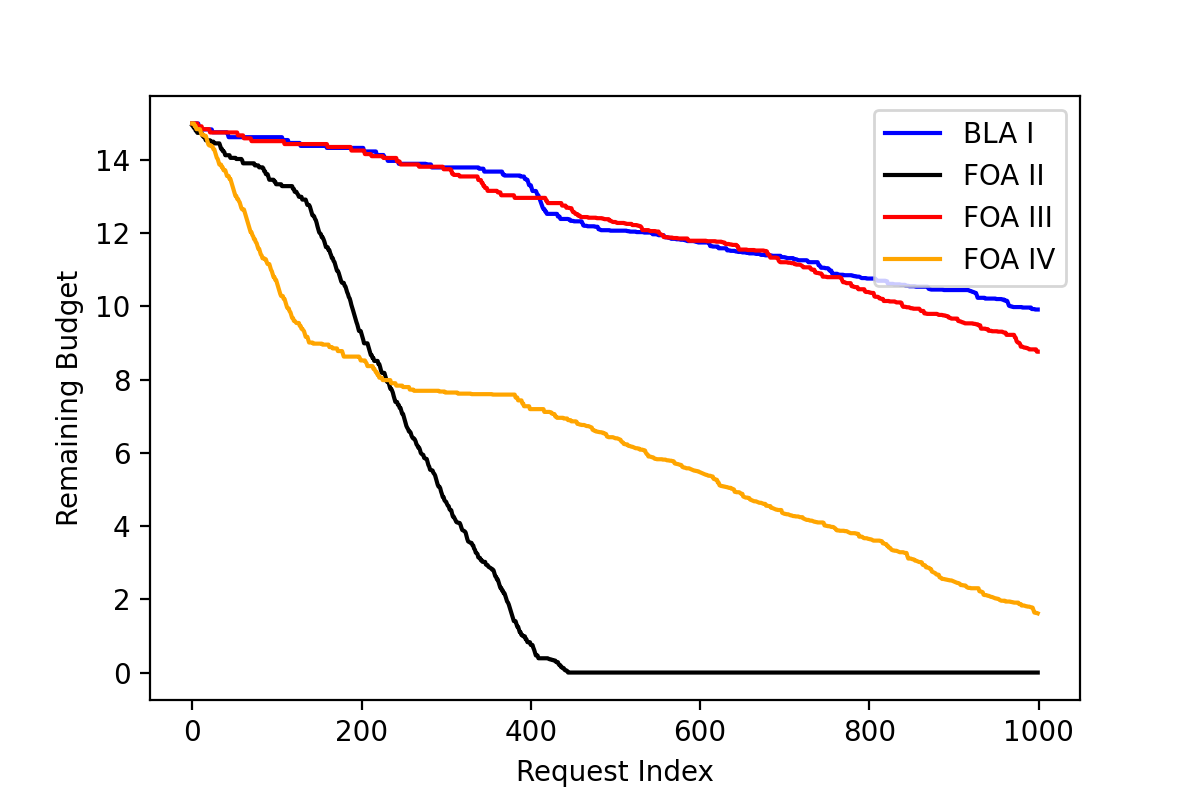}}\hspace*{-0.9em} 
\subfigure{
    \includegraphics[width=4.5cm]{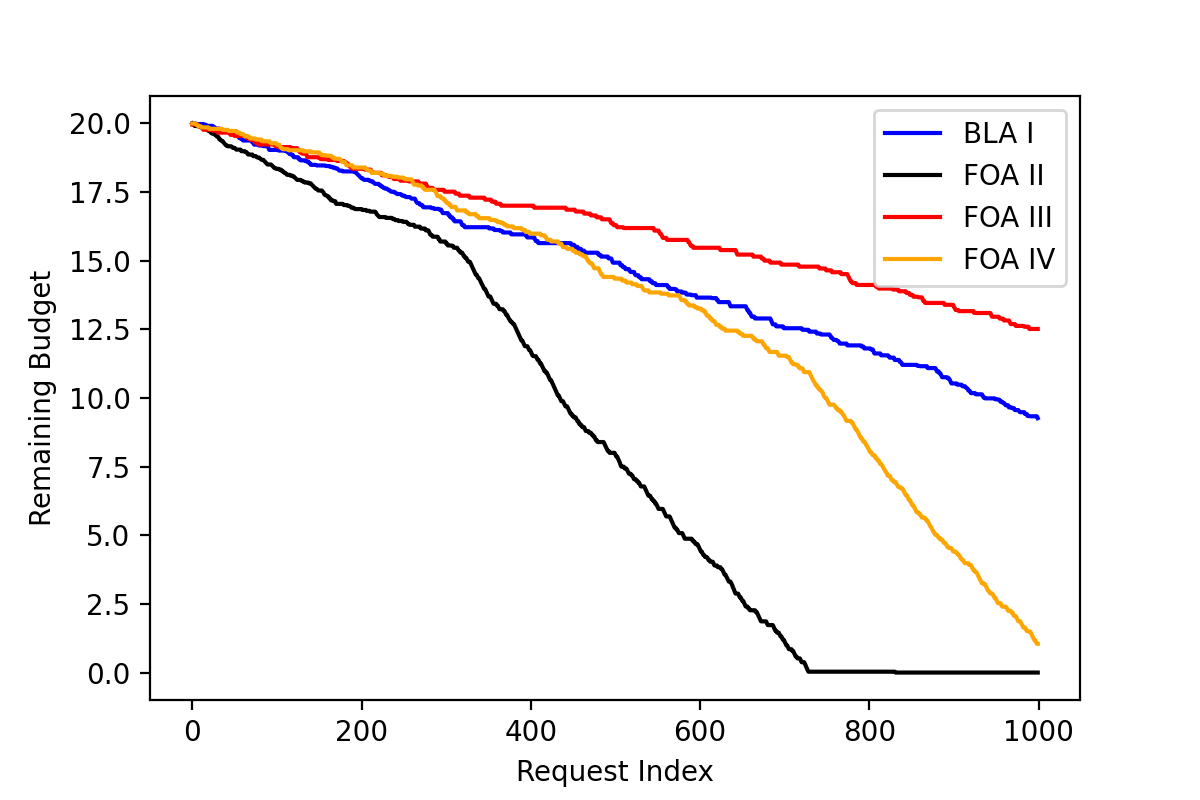}}\hspace*{-0.9em}
    \\
    \vspace*{-0.9em}
\subfigure[(a) Budget=5K]{
    \includegraphics[width=4.5cm]{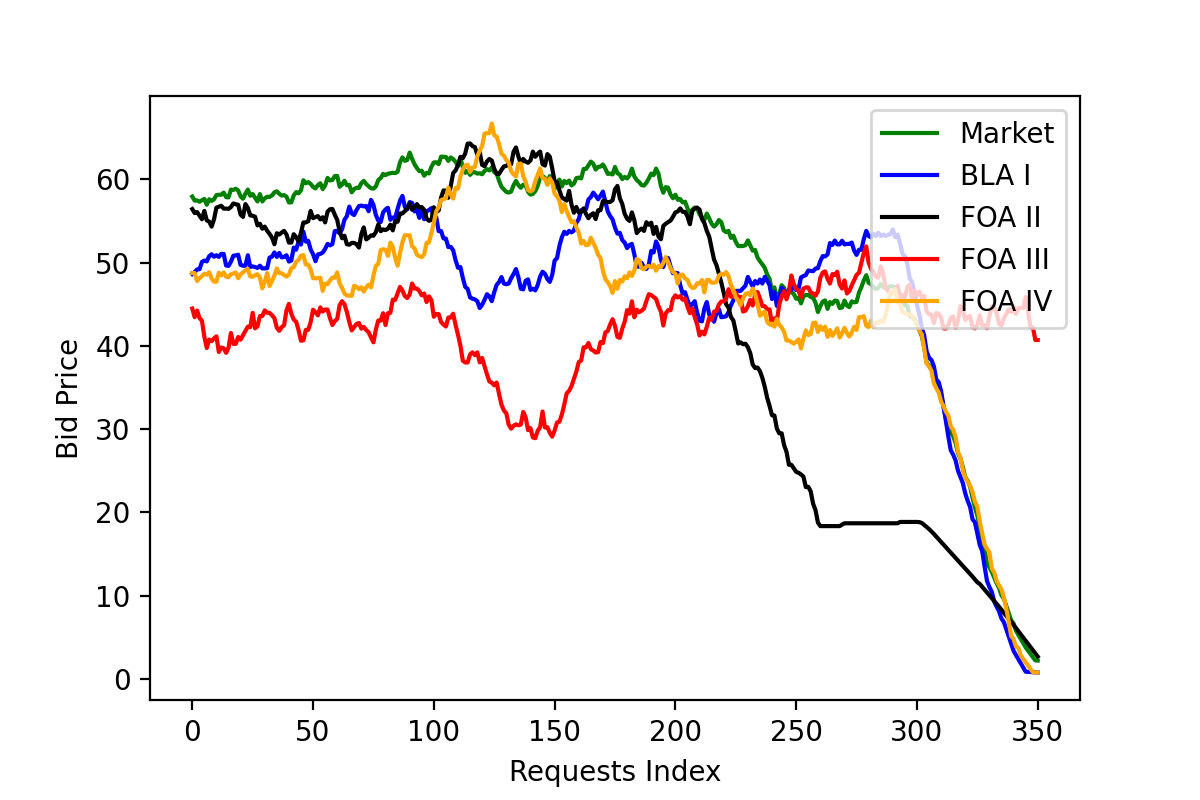}}\hspace*{-0.9em} 
\subfigure[(b) Budget=10K]{
    \includegraphics[width=4.5cm]{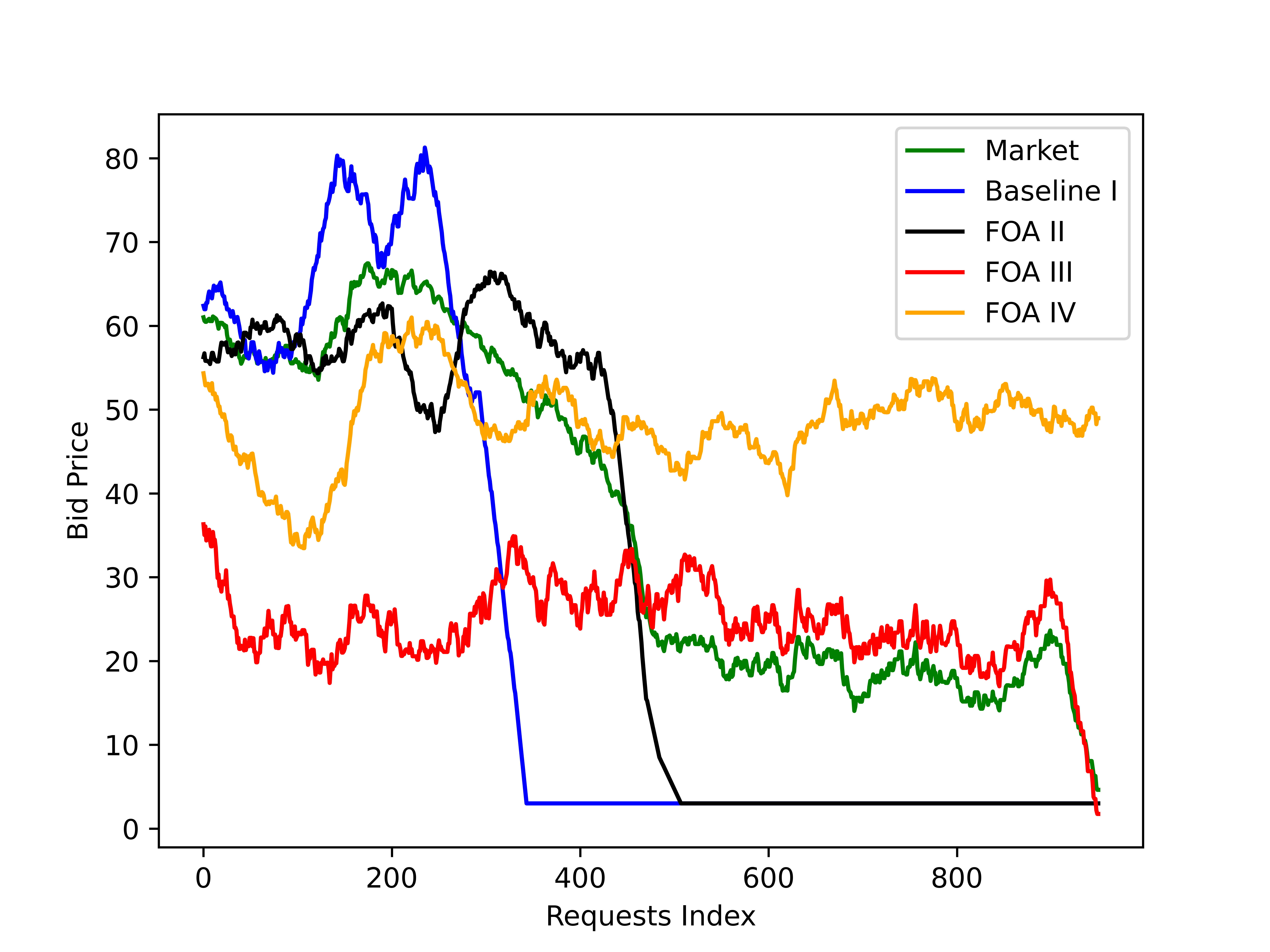}}\hspace*{-0.9em} 
\subfigure[(c) Budget=15K]{
    \includegraphics[width=4.5cm]{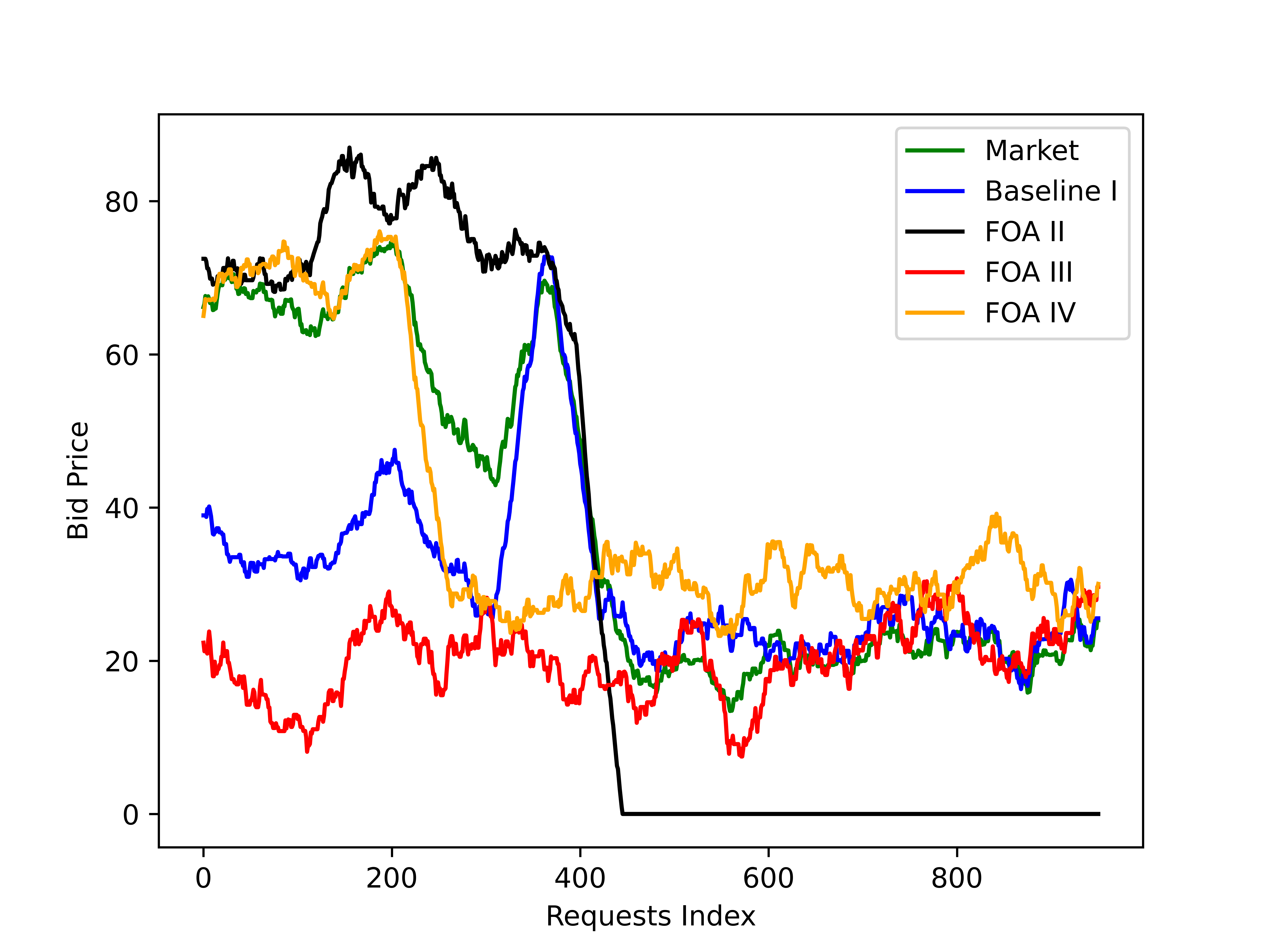}}\hspace*{-0.9em}
\subfigure[(d) Budget=20K]{
    \includegraphics[width=4.5cm]{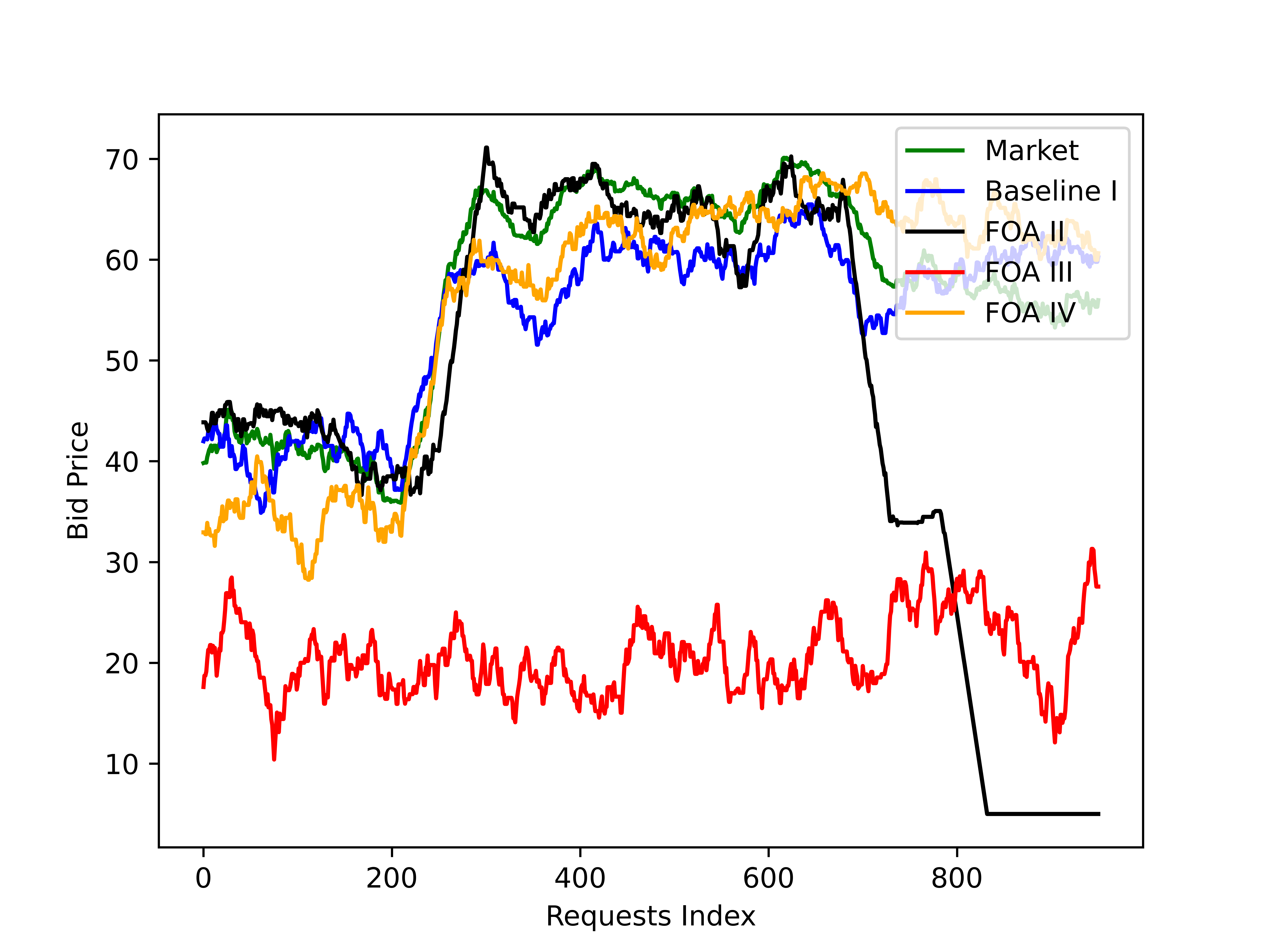}}\hspace*{-0.9em}
\caption{Unbiased algorithm comparison results: Budget Consumption (first row) and Distribution of the bid price and winning (second row).}
\label{results_unbiased}
\end{figure*}
 The graphical comparison results are shown in Fig \ref{results_biased} and Fig \ref{results_unbiased}, with the first row displaying the budget consumption and the second row displaying the distribution of bid price and winning.
In the second row, we use moving average windows to smooth the high-frequency data. Fig \ref{results_biased} results shows that the decline rate of remaining budget for \textbf{FOA III} is significantly greater than others over different budgets. Consequently, for biased winning probability when budget is limited (under \$750K in our experiment), FOA III is crafty and learns to bid less at the begining until other agents have exhausted their budgets and then get plenty of opportunities to win. The bidding price subfigures in Fig \ref{results_biased} also illustrate the bidding strategy learned by FOA III. \\
\indent Table \ref{biased_table} and \ref{unbiased_table} indicate the average Win Rate and Win Surplus Over for 10 campaigns for biased and unbiased Winning Probability respectively with different budgets. 
 Win surplus indicates the difference between the first price(winning price) and the second price(market price). The higher win surplus demonstrates the agent can help users save more money while guaranteeing the chance of winning.\\
\indent According to Table \ref{biased_table}, for diverse budgets 250K, 500K, 750K, the FOA III gets the average win surplus of $48.19\$, 46.68\$, 81.19\$$ respectively. Function agents II and IV perform better than baseline agent I by capturing more data from the whole campaign. Therefore, the knowledge of the prior biased distribution is good guidance for the agent to bid. However, any strategy will become useless when each agent has adequate budget(more than \$1000K in our case). Becasue the constraint of equation (\ref{KPI_based}) is no longer strict and agents have more freedom to bid ,so that the impact of Lagrange multiplier $\lambda$ in helping agents make optimal decisions becomes slighter. As a result, the baseline agent wins at $34.72\%$ at the budget 1250K. \\
\indent Table \ref{unbiased_table} shows the superiority of \textbf{FOA IV} with unbiased winning probability. Still, for limited budget (\$5K-\$15K), FOA IV which considers $\lambda$ as reward has the robust performance with the highest win rate and win surplus among the four. Due to equation (\ref{10}), the unbiased probability allows the agent to get feedback from the whole market and serve as its reward to make optimal decisions from a global perspective. The second row of Fig \ref{results_unbiased} also intutively shows the bidding price of FOA IV is generally higher than baseline agent I's. However, as the budget increases, the same problems arise as biased campaigns. Table \ref{unbiased_table} shows for budget 25K, the baseline agent has slightly greater performance than FOA II and FOA IV and is significantly more powerful than FOA III.
\section{Conclusion}
In this paper, we design the multi-agent reinforcement learning algorithm combined with functional optimization. We created Lagrange multiplier-based agents to take use of functional optimization's capabilities. The results on simulated campaigns demonstrate the effectiveness of functional agents in RTB.
In case of limited budget, some functional agents perform better than the baseline agent and have robust performance when the budget varies.
The functional agent (as action) bids towards winning data in the situation of a biased winning probability model. When the unbiased winning probability model is used, the functional agent (as reward) takes the campaign's global information and uses the unbiased reward to make the best decisions.
\\
In future work, we can implement the proposed framework to real data to test the efficiency of the use of functional optimization with reinforcement learning. We can also try to develop more strategies which can be stable for the performance of the functional optimization agents on a different pattern of campaigns.
\bibliographystyle{ieeetr}
\bibliography{Reference}
\end{document}